\documentclass[runningheads]{llncs}

\usepackage{accv}

\usepackage{accvabbrv}

\usepackage{graphicx}
\usepackage{booktabs}

\usepackage{algorithm}
\usepackage{algpseudocode}

\usepackage{pifont}
\newcommand{\cmark}{\ding{51}}
\newcommand{\xmark}{\ding{55}}
\usepackage{multirow}

\usepackage[accsupp]{axessibility}  % Improves PDF readability for those with disabilities.

\usepackage{hyperref}

\usepackage{orcidlink}

\makeatletter
\newcommand{\printfnsymbol}[1]{%
        \textsuperscript{\@fnsymbol{#1}}%
}
\makeatother

\usepackage{bibunits}
\defaultbibliographystyle{splncs04}

\begin{document}

\title{Efficient Scene-Distribution-Aware Second-Order Optimization for 3D Gaussian Splatting} 

\titlerunning{Efficient Second-Order Optimization for 3DGS}

\author{
Hyunseung Son\inst{1}\orcidlink{0009-0009-1239-0492} \and
Young Sun Choi\inst{1}\orcidlink{0009-0001-9836-4245} \and
Youngsik Yun\inst{1}\orcidlink{0000-0003-4398-7856} \and
Kibeom Hong\inst{2}\thanks{Corresponding author.}\orcidlink{0000-0003-2366-6318} \and
Youngjung Uh\inst{1}\printfnsymbol{1}\orcidlink{0000-0001-8173-3334}
}

\authorrunning{H.~Son et al.}

\institute{
Department of Artificial Intelligence, Yonsei University, Seoul, Republic of Korea\\
\email{\{ghfod0917,youngsun.choi,bbangsik,yj.uh\}@yonsei.ac.kr}
\and
Sookmyung Women’s University, Seoul, Republic of Korea\\
\email{kb.hong@sookmyung.ac.kr}
}

\maketitle

\begin{abstract}

3D Gaussian Splatting (3DGS) optimization updates the Gaussian parameters one training view at a time, tens of thousands of times.
Unlike such single-view updates, multi-view batches reduce dependence on individual training views, and second-order optimization reduces the number of updates with larger, structured parameter changes.
Because fewer updates leave fewer chances to correct a change that fits only the views used in one update,
scene coverage becomes more important for each second-order update.
Multi-view second-order updates, however, are expensive.
Pixel sampling reduces this cost, but two issues remain.
First, selecting views without considering overlap leaves some views redundant. To address this, we propose a view selection strategy that reduces this redundancy.
Second, existing work processes the sampled pixels of each view in a separate pass and recomputes alpha compositing for these pixels from scratch.
We introduce two implementation strategies that reduce both overheads by processing the sampled pixels of multiple views in joint passes and reusing the transmittance and accumulated color cached during rendering.
On three datasets, our method reaches higher PSNR than the sampled second-order baseline in less than half its optimization time, with competitive SSIM and LPIPS.

\end{abstract}
\section{Introduction}
\label{sec:intro}
3D Gaussian Splatting (3DGS)~\cite{kerbl2023gaussians} renders novel views photorealistically in real time, but optimizing it is slow.
Converging with Adam takes 7.6 minutes for 30,000 single-view updates even from a sparse SfM point cloud, and still 6.8 minutes for 10,000 fine-tuning updates from a dense pretrained 3DGS model, on the Mip-NeRF~360 scenes~\cite{barron2022mipnerf360} with a single GPU (\cref{sec:exp}).
Even recent feed-forward pipelines~\cite{jiang2025anysplat,an2026c3g} predict Gaussians in one pass and then refine them with roughly 1,000 to 3,000 updates.
Therefore, faster optimization benefits training from scratch, fine-tuning, and feed-forward refinement alike.

Existing methods either make each update cheaper by accelerating rendering or backpropagation~\cite{durvasula2023distwar,taming3dgs,flashgs,speedysplat}, or reach a given quality in fewer updates.
Building each update from a batch of views lets multi-view batching methods~\cite{mvgs,grendel,efficientmultiview3dgs} reach a given quality in fewer updates.
Scaling each update by the curvature of the loss lets second-order methods~\cite{3dgslm,lmrs,3dgs2,3dgs2tr} need fewer updates still.
With fewer updates, an update that lowers the loss on its own views but raises it on other views is corrected by fewer later updates.
Each second-order update therefore needs its views to cover the scene broadly.
At a nearly fixed number of sampled pixels, more views per batch give a better second-order update (\cref{sec:method_overview}).
However, combining multi-view batches with second-order updates is expensive because every view in a batch must be rendered, differentiated, and added to the second-order system.

Among second-order methods, LM-RS~\cite{lmrs} solves jointly for one update of all the Gaussian parameters it optimizes from all views of a batch.
LM-RS forms the system from a sampled subset of pixels of each view, and the system then takes less time and less memory to solve than with all pixels.
However, LM-RS assumes Lambertian surfaces and optimizes one color per Gaussian, while standard 3DGS models let color vary with the viewing direction through additional spherical-harmonics (SH) coefficients.
On a standard 3DGS model, LM-RS leaves the additional SH coefficients unoptimized.
We keep the pixel sampling of LM-RS but also optimize the additional SH coefficients.
LM-RS with the additional SH coefficients reaches a PSNR close to that of 30,000 single-view Adam updates with 300 updates of 64 views each.
Although 300 updates are 100 times fewer than 30,000, the 300 LM-RS updates take 8.2 minutes, while the 30,000 Adam updates take 7.6 minutes (\cref{sec:exp_results}).
Two issues remain, one in how much each update improves reconstruction quality and one in how long each update takes.

First, LM-RS selects the views of each batch by K-Means clustering of camera poses.
Selecting views independently across clusters lets two nearly identical views from neighboring clusters enter the same batch.
The two overlapping views leave other regions of the scene under-represented in the update built from the batch.
We introduce \emph{Scene-Distributional View Sampling}, which places the views of each batch evenly along the camera trajectory, replaces any view that overlaps with those already in the batch by a less similar view nearby, and determines all batches before optimization.
The strategy reduces redundancy within a batch without extra rendering or backward passes.

Second, LM-RS runs a separate pass over each view of a batch to accumulate over the sampled pixels for solving the Levenberg--Marquardt (LM) system.
LM-RS also recomputes each pixel’s front-to-back alpha compositing from the first Gaussian to compute the preconditioner.
We introduce two implementation strategies that shorten the accumulation and the preconditioning without modifying the update rule.
\emph{Tile-Grouped Multi-View Accumulation} lets one accumulation pass cover several views by packing their tiles into one layout.
\emph{Bucket-Prefix Jacobi Preconditioning} caches the transmittance and accumulated color at a few Gaussians per ray during rendering, so the preconditioner computation starts from them in parallel instead of from the first Gaussian.
Together, the two strategies shorten each multi-view sampled LM update.

On three datasets, our method reaches higher reconstruction quality than the sampled second-order baseline in less than half its optimization time, and higher PSNR than single-view Adam in less time.
The results show that our second-order optimization is a practical choice for 3DGS, and not only a way to reduce the number of updates.
The gain holds for optimization starting from either a SfM point cloud or a pretrained 3DGS model.

Our contributions are summarized as follows:
\begin{itemize}
    \item \emph{Scene-Distributional View Sampling}, which selects views by camera pose and outperforms random and K-Means selection in PSNR at equal batch size.
    \item \emph{Tile-Grouped Multi-View Accumulation} and \emph{Bucket-Prefix Jacobi Preconditioning},  strategies that make each sampled LM update $2\times$ faster.
    \item Higher PSNR than single-view Adam and second-order baselines in less time, both from sparse SfM point clouds and from dense pretrained 3DGS models.
\end{itemize}

\section{Related Work}
\label{sec:rel}

\subsection{Training Acceleration for 3D Gaussian Splatting}
\label{sec:rel_accel}
Existing training acceleration methods shorten each update by limiting the number of Gaussians or by speeding up rasterization and backpropagation kernels.
% lightgaussian은 training 가속아니라 뺏어요.
Mini-Splatting~\cite{minisplatting} and FastGS~\cite{fastgs} remove Gaussians during training.
3DGS-MCMC~\cite{3dgsmcmc} relocates Gaussians under a user-set maximum number.
FlashGS~\cite{flashgs} and Speedy-Splat~\cite{speedysplat} speed up rasterization, and DISTWAR~\cite{durvasula2023distwar} and Taming3DGS~\cite{taming3dgs} speed up backpropagation.
These acceleration methods leave the number of updates to reach a given quality unchanged.
To reduce the number of updates instead, other methods build each update from several views and the curvature of the loss, but building such an update takes longer than summing the gradients from those views.
Our method shortens building such an update.

\subsection{Batching and View Selection in 3DGS Optimization}
\label{sec:rel_batching}
Multi-view batching builds one update from a multi-view batch and reaches a given quality in fewer updates than single-view training.
MVGS~\cite{mvgs} sums the gradients over the batch so that one update covers the Gaussians of all its views.
Efficient Multi-View 3DGS~\cite{efficientmultiview3dgs} shows that the gradient estimate from a multi-view batch has lower variance across training views than the gradient from one view, and renders only a subset of the pixels of each view to keep the update fast.
Grendel~\cite{grendel} spreads the views of a multi-view batch across several GPUs to train on them at once.
MVGS and Grendel draw the views of a batch uniformly at random.
Subset-selection methods~\cite{gonzalez1985clustering,kulesza2012dpp,chen2010kernelherding} pick samples that represent or cover a dataset under fixed descriptors.
LM-RS~\cite{lmrs} clusters camera poses with K-Means and selects one random view from each cluster at every update.
Active methods~\cite{activenerf,fisherrf,popgs} instead select views to add to the training set by scoring each candidate with the current reconstruction.

Random batching and cluster-based selection pick each view without checking its overlap with the other views of the batch.
Random batching does not control which views share a batch.
Cluster-based selection draws one view from each cluster independently.
Active methods require a rendering or backward pass for each candidate view.
Our selection needs only camera poses and is computed before optimization, with no extra rendering or backward pass.
At the same batch size, our selection checks overlap among the selected views and replaces an overlapping view with a less similar view nearby along the camera trajectory.

\subsection{Second-order 3DGS Methods}
\label{sec:rel_second_order}
Second-order optimization for 3DGS in its full formulation solves one system that couples all Gaussian parameters using all views of a multi-view batch. However, because this full system is expensive, some methods solve separate systems over subsets of views or parameters.
3DGS-LM~\cite{3dgslm} averages updates over view subsets, and 3DGS$^2$~\cite{3dgs2} solves a local system per parameter type.
3DGS$^2$-TR~\cite{3dgs2tr} keeps only the diagonal of the system and bounds each update with a parameter-wise trust region.
TurboGS~\cite{turbogs} solves a local system per Gaussian from sampled pixels and updates the SH coefficients separately with Adam.
Local parameter systems shorten each update and leave the cross terms between uncoupled parameter groups to later updates, while a full system retains these cross terms within each update and takes longer per update.
In contrast, LM-RS~\cite{lmrs} keeps one coupled system and shortens each update in two ways: it assumes Lambertian surfaces and leaves the additional SH coefficients out of the system, and it forms the system from sampled pixels.
Each LM-RS update also runs a separate pass for each view of the batch and recomputes each pixel's alpha compositing from the first Gaussian for the preconditioner.
Our method keeps the full system with the additional SH coefficients, selects multi-view batches with less redundancy, and shortens the backward and preconditioning passes whose time grows with the number of views in the batch.
\section{Method}
\label{sec:method}

\subsection{Preliminaries}
\label{sec:preliminaries}

\noindent\textbf{3D Gaussian Splatting.} 
3DGS~\cite{kerbl2023gaussians} represents a scene as a set of anisotropic 3D Gaussian primitives, denoted  $\boldsymbol{\theta}\in\mathbb{R}^P$, where $P$ is the total number of parameters.
Each Gaussian $g\in\boldsymbol{\theta}$ is parameterized by an optimizable center position $\boldsymbol{\mu}_g\in\mathbb{R}^3$, a scale $\boldsymbol{s}_g\in\mathbb{R}^3$, a rotation given by a unit quaternion $\boldsymbol{q}_g\in\mathbb{R}^4$, an opacity $o_g\in\mathbb{R}$, and spherical-harmonics coefficients %\kb{$\boldsymbol{c}_g$}
that produce view-dependent color. The scale and quaternion are assembled into a scaling matrix $\boldsymbol{S}_g\in\mathbb{R}^{3\times3}$ and a rotation matrix $\boldsymbol{R}_g\in\mathbb{R}^{3\times3}$, which together define the covariance $\boldsymbol{\Sigma}_g\in\mathbb{R}^{3\times3}$. The spatial influence of Gaussian $g$ at a 3D position $\boldsymbol{x}\in\mathbb{R}^3$ is:
\begin{equation*}
    \mathcal{G}_g(\boldsymbol{x})
    =
    \exp\!\left[
    -\frac{1}{2}
    (\boldsymbol{x}-\boldsymbol{\mu}_g)^{\top}
    \boldsymbol{\Sigma}_g^{-1}
    (\boldsymbol{x}-\boldsymbol{\mu}_g)
    \right],
    \qquad
    \boldsymbol{\Sigma}_g
    =
    \boldsymbol{R}_g
    \boldsymbol{S}_g
    \boldsymbol{S}_g^{\top}
    \boldsymbol{R}_g^{\top}.
    \label{eq:prelim_gaussian}
\end{equation*}

To render an image, 3D Gaussians are projected onto the image plane, sorted in depth order, and alpha-composited front-to-back. 
The equation for rendering the pixel for ray $i$ is as follows:

\begin{equation*}
    \boldsymbol{C}_{i}
    =
    \sum_{n=1}^{N_i}
    T_{i,n}\,\alpha_{i,n}\,\boldsymbol{c}_{i,n},
    \qquad
    T_{i,n}
    =
    \prod_{m<n}(1-\alpha_{i,m}),
    \label{eq:prelim_compositing}
\end{equation*}

where $\boldsymbol{C}_{i}\in\mathbb{R}^3$ is the rendered color of ray $i$ and
$N_i$ is the number of Gaussians composited along ray $i$.
$\boldsymbol{c}_{i,n}\in\mathbb{R}^3$ and $\alpha_{i,n}\in\mathbb{R}$ are
the color and alpha of the $n$-th composited Gaussian. 
$T_{i,n}\in\mathbb{R}$ is the accumulated transmittance by the Gaussians in front of the $n$-th Gaussian.

3DGS is trained by minimizing the residual $r_{i,k}$ between the rendering and ground truth with least-squares and D-SSIM losses, where $i=1,\dots,M$ indexes the $M$ rays across all training views and $k$ is the color channel. 
The residual and least-squares loss are expressed as:

\begin{equation}
    r_{i,k}(\boldsymbol{\theta})
    =
    C_{i,k}(\boldsymbol{\theta})-C^{\mathrm{gt}}_{i,k},
    \qquad
    \mathcal{L}(\boldsymbol{\theta})
    =
    \frac{1}{2}
    \sum_{i=1}^{M}\sum_{k=1}^{3}
    r_{i,k}(\boldsymbol{\theta})^2
    =
    \frac{1}{2}
    \|\boldsymbol{r}(\boldsymbol{\theta})\|_2^2 ,
    \label{eq:prelim_objective}
\end{equation}

where $C_{i,k}$ is the $k$-th channel of the rendered color $\boldsymbol{C}_i$
and $C^{\mathrm{gt}}_{i,k}$ is the corresponding ground-truth value.
$\boldsymbol{r}(\boldsymbol{\theta})\in\mathbb{R}^{3M}$ is the vector of all residuals $r_{i,k}$ over the $M$ rays and 3 color channels.

\noindent\textbf{Levenberg--Marquardt optimization.}
Second-order 3DGS optimizers use a Gauss--Newton or LM approximation to compute a curvature-aware parameter update $\boldsymbol{\delta}\in\mathbb{R}^P$. 
Let $\boldsymbol{J}=\partial\boldsymbol{r}/\partial\boldsymbol{\theta}\in\mathbb{R}^{3M\times P}$ be the Jacobian of the residual vector $\boldsymbol{r}$ with respect to the Gaussian parameters $\boldsymbol{\theta}$.
Hence, the LM update $\boldsymbol{\delta}$ is obtained by solving:
\begin{equation*}
    \left(
    \boldsymbol{J}^{\top}\boldsymbol{J}
    +
    \lambda\boldsymbol{I}
    \right)
    \boldsymbol{\delta}
    =
    -\boldsymbol{J}^{\top}\boldsymbol{r},
    \label{eq:full_lm_system}
\end{equation*}
where $\lambda$ is the damping coefficient. 
The full system is infeasible to form in 3DGS, because $\boldsymbol{r}$ contains every pixel-channel error of all training views and $\boldsymbol{\theta}$ contains the parameters of millions of Gaussians.
Existing second-order methods therefore either solve smaller systems over subsets of views or parameters, or, as in LM-RS~\cite{lmrs}, form the coupled system from sampled pixels and solve it with a matrix-free preconditioned conjugate gradient (PCG) solver, which needs only products with the Jacobian and its transpose.

\noindent\textbf{Sampled LM Formulation.}
A sampled LM step approximates the full system by drawing a smaller set of rays and assembling the update from their residuals. 
The sampling of rays is expressed as a distribution $q$ over rays. Thus $q(i)$ is the sampling probability of ray $i$, where the ray index $i$ is defined for every view, tile, and pixel, giving $M$ rays in total.
This distribution determines which views are sampled, and how many rays are drawn from each. 
The least-squares objective in \cref{eq:prelim_objective} can be
rewritten as an expectation over rays sampled from the distribution $q$:

\begin{equation*}
    \mathcal{L}(\boldsymbol{\theta})
    =
    \frac{1}{2}
    \sum_{i=1}^{M}\sum_{k=1}^{3}
    r_{i,k}(\boldsymbol{\theta})^2
    =
    \frac{1}{2}
    \mathbb{E}_{i\sim q}
    \left[
    \frac{1}{q(i)}\sum_{k=1}^{3} r_{i,k}(\boldsymbol{\theta})^2
    \right].
    \label{eq:distributional_objective}
\end{equation*}

In practice, a sampled LM step approximates this objective using a small subset $\mathcal{S}$ of the $M$ rays.
Let $\widehat{\boldsymbol{J}}$ and $\boldsymbol{r}_{\mathcal{S}}$ denote the Jacobian and residual over the
sampled rays $\mathcal{S}$.
The resulting sampled LM system is

\begin{equation}
    (
    \widehat{\boldsymbol{J}}^{\top}\widehat{\boldsymbol{J}}
    +
    \lambda\boldsymbol{I}
    )
    \boldsymbol{\delta}
    =
    -\,\widehat{\boldsymbol{J}}^{\top}\boldsymbol{r}_{\mathcal{S}} .
    \label{eq:sampled_lm_system}
\end{equation}

Since $q$ determines $\mathcal{S}$, it also determines which rows enter $\widehat{\boldsymbol{J}}$.
Views that observe nearly the same part of the scene contribute nearly identical rows, which add little new information to $\widehat{\boldsymbol{J}}^{\top}\widehat{\boldsymbol{J}}$, and Gaussians seen by no selected view receive no update from this step.
Our method selects the views of each batch so that their camera poses are distributed more like those of all training views.
% As $q$ determines the sampled set $\mathcal{S}$, and hence
% $\widehat{\boldsymbol{J}}$ and $\boldsymbol{r}_{\mathcal{S}}$, 
% a poorly chosen distribution lets redundant views dominate the update, so the step reflects only a narrow part of the scene. 
% Our method addresses this by constructing a view sampling distribution that covers the scene more evenly.

\begin{figure*}[t]
\centering
\includegraphics[width=\textwidth]{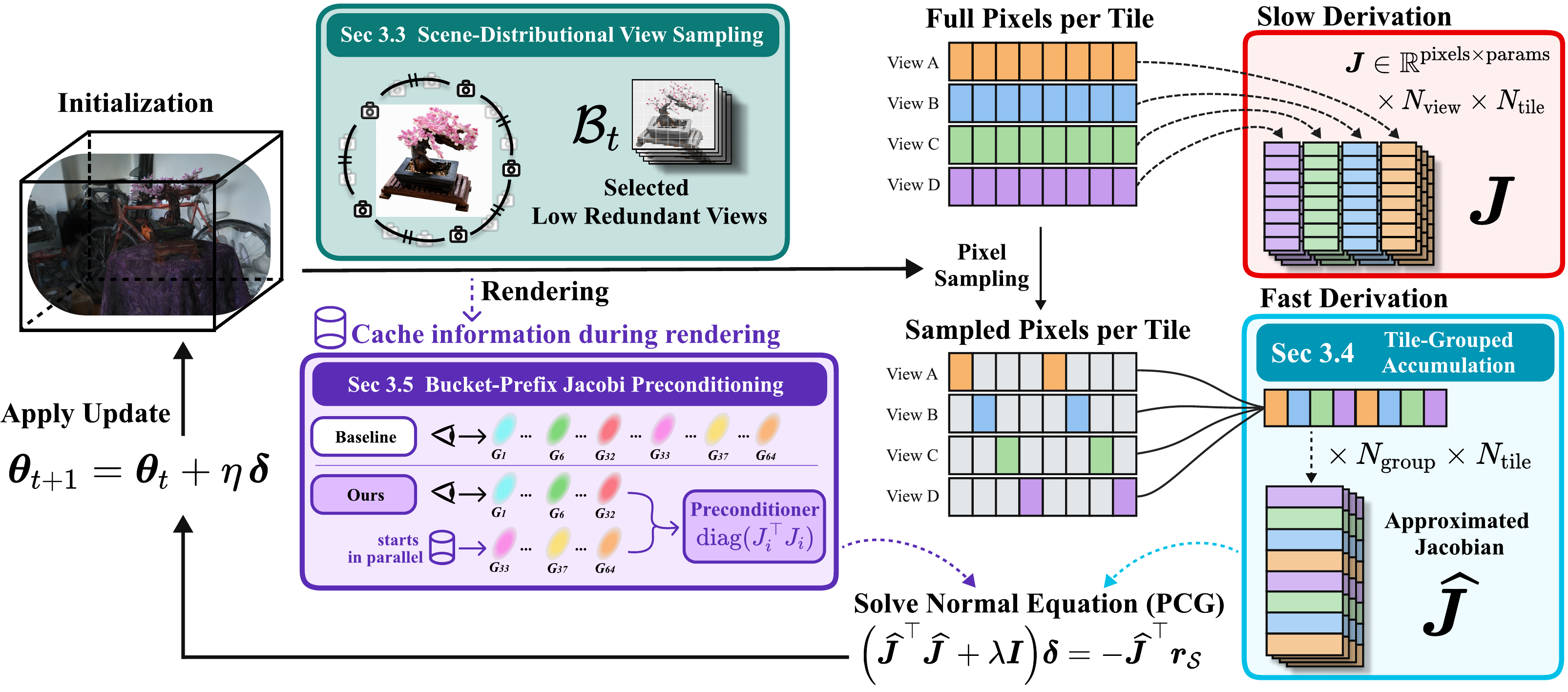}
\caption{
\textbf{Overview of one update of our method.}
From Gaussians initialized by a sparse SfM point cloud or a dense pretrained 3DGS model, Scene-Distributional View Sampling selects training views with low redundancy, forming the multi-view batch $\mathcal{B}_t$. 
The views are rendered, and pixels are sampled per tile so that their rays provide residuals.
Tile-Grouped Multi-View Accumulation packs the sampled rays from view-local tiles into a single grouped layout and derives $\hat{\boldsymbol{J}}$ for all views jointly.
Bucket-Prefix Jacobi Preconditioning caches the transmittance and accumulated color at a few Gaussians per ray during rendering, so the preconditioner computation starts from them in parallel.
$N_{\mathrm{view}}$, $N_{\mathrm{tile}}$, and $N_{\mathrm{group}}$ denote the numbers of selected views, tiles per view, and grouped tiles.}

\label{fig:method_overview}
\end{figure*}
\begin{table}[t]
\centering
\caption{
\textbf{With nearly the same number of sampled pixels, more views improve PSNR but increase time and memory.}
We use Bonsai from Mip-NeRF 360~\cite{barron2022mipnerf360}, whose 255 training views let the number of views double from 32 to 64, 128, and all 255, and run the same sampled LM solver in each setting.
Each additional view requires another rendering pass, and its per-view rendering results must be cached for optimization.
Rays is the total number of sampled pixels in millions, Time is the optimization time, and Mem. is the peak memory.
Best PSNR is in \textbf{bold}.
}

\label{tab:view_budget}
\footnotesize
\setlength{\tabcolsep}{3pt}
\begin{tabular*}{\linewidth}{@{\extracolsep{\fill}}lcccccc@{}}
\toprule
Setting
& Views
& Pixel/Tile
& Rays (M)
& PSNR$\uparrow$
& Time (s)$\downarrow$
& Mem. (GB)$\downarrow$ \\
\midrule
LM-RS Default
& 32
& 32
& 6.52
& 30.36
& 321.2
& 12.48 \\
Broader Views
& 64
& 16
& 6.52
& 30.80
& 578.1
& 18.33 \\
Broader Views+
& 128
& 8
& 6.52
& 31.04
& 1054.2
& 30.67 \\
Full Views
& 255
& 4
& 6.50
& \textbf{31.23}
& 1916.4
& 55.06 \\
\bottomrule
\end{tabular*}
\end{table}

\subsection{Overview}
\label{sec:method_overview}

One update of our method is illustrated in \cref{fig:method_overview}.
Each sampled pixel defines one ray, and the sampled LM system is formed from the rays of all views in the batch.
Since an update is built from only a small set of rays, its quality depends not only on how many rays are sampled but also on which views they are drawn from.
At a fixed number of rays, drawing fewer rays from more views gives higher PSNR than drawing more rays from fewer views (\cref{tab:view_budget}).
However, broad view coverage is expensive in practice, because the backward time of each second-order update grows with the number of views in the batch.
Our method therefore addresses two challenges: constructing a multi-view batch that represents all training views, and shortening the backward pass over a batch.

Our method extends the sampled LM solver with three components.
\emph{Scene-Distributional View Sampling} (\cref{sec:scene_distribution_sampling}) selects a batch that covers the training views more evenly at a fixed number of views and reduces redundancy among the views in the same batch.
\emph{Tile-Grouped Multi-View Accumulation} (\cref{sec:tile_grouped_backward}) accumulates the Jacobian products over the rays of multiple views in one pass over a single grouped layout rather than view by view.
\emph{Bucket-Prefix Jacobi Preconditioning} (\cref{sec:bucket_prefix}) reuses cached alpha compositing values from rendering to compute the preconditioner efficiently.

\subsection{Scene-Distributional View Sampling}
\label{sec:scene_distribution_sampling}
Scene-Distributional View Sampling selects the $K$ views of each batch from camera poses.
Rays from the same view share one camera position and lie within its field of view, so we compare views by camera position and viewing direction rather than by the rays they induce, which are expensive to estimate per view.
The batch is built in three steps: ordering the views by camera position, placing evenly spaced anchors along this order, and refining each anchor to reduce redundancy among the views already in the batch.

\noindent\textbf{Pose-geodesic camera order.}
The input image order need not follow the camera placement, so we first order the cameras by position.
For each training view $v$, let $\boldsymbol{c}_v$ be its camera center.
We normalize the camera centers, build a symmetric $k$-nearest-neighbor graph on them, take the two cameras farthest apart on this graph, and order all cameras by their graph shortest-path distance from one of the two.
This gives the pose-geodesic order $\pi=(\pi_0,\ldots,\pi_{N-1})$, where $\pi_i$ is the training view at index $i$; nearby indices correspond to nearby camera poses, so the order follows the camera trajectory even when the input image list is shuffled.

\noindent\textbf{Uniform view anchors.}
We place the $K$ views at evenly spaced positions along the pose-geodesic order rather than clustering them in one region, and shift these positions by one index at every LM step so that successive batches cover all views over time.
The anchor for the $j$-th view at LM step $t$ is
\begin{equation*}
    a_{t,j} = \left(\operatorname{round}\left(\frac{j(N-1)}{K-1}\right) + t\right) \bmod N, \qquad j=0,\ldots,K-1,
    \label{eq:phase_slots}
\end{equation*}
an index into $\pi$, not yet the final selected view.
Because uniform spacing in index does not guarantee uniform coverage in camera position or viewing direction, we refine each anchor within the same step.

\noindent\textbf{Local redundancy-aware refinement.}
The anchors spread the views along the order, but the index is only a rough proxy for camera position and viewing direction, so two anchors can still land on views that observe nearly the same thing.
We therefore refine each anchor within a window of $w$ indices around it, comparing views by a pose-direction descriptor
\begin{equation*}
    \boldsymbol{\phi}_i = \left[\boldsymbol{d}_i,\; \frac{(\boldsymbol{c}_{\pi_i}-\bar{\boldsymbol{c}})\times\boldsymbol{d}_i}{\rho}\right], \qquad \hat{\boldsymbol{\phi}}_i = \frac{\boldsymbol{\phi}_i}{\|\boldsymbol{\phi}_i\|_2},
    \label{eq:pose_direction_feature}
\end{equation*}
where $\boldsymbol{d}_i$ is the viewing direction of view $\pi_i$, $\bar{\boldsymbol{c}}$ is the mean camera center, and $\rho$ is the mean distance of the camera centers from $\bar{\boldsymbol{c}}$.
Within each window we select the candidate whose descriptor has the lowest maximum cosine similarity to the views already chosen for the batch.
After all anchors are refined, the selected views form the batch $\mathcal{B}_t$.

The sampler changes only which views the pixels are sampled from; pixel sampling, accumulation, damping, and the sampled LM solve are unchanged from LM-RS.
On Mip-NeRF 360, the 64-view batches of our sampler have a lower Coverage Distance to all training views in camera position and viewing direction than the K-Means batches of LM-RS (\cref{tab:mip360_sampling_ablation}).
Although the sampler does not control which rays are sampled, its batches also give a ray distribution closer to that of all training views, as shown in the supplementary material.

\begin{table}[t]
\centering
\caption{
\textbf{View selection on Mip-NeRF 360.}
All methods use 64 views, sample 32 pixels per tile, and perform 300 sampled LM updates from the same SfM initialization.
Coverage Distance compares the camera positions and viewing directions of the selected views with those of all training views over 100 batches.
All values are averaged over the nine scenes.
Best Coverage Distance and PSNR are in \textbf{bold}.
}
\label{tab:mip360_sampling_ablation}

\footnotesize
\setlength{\tabcolsep}{4pt}

\begin{tabular*}{\linewidth}{@{\extracolsep{\fill}}lccccc@{}}
\toprule
Sampling
& Views
& Pixel/Tile
& Coverage Distance$\downarrow$
& PSNR$\uparrow$
& SSIM$\uparrow$ \\
\midrule
Random
& 64
& 32
& 0.003588
& 25.661
& 0.6789 \\
K-Means (LM-RS)
& 64
& 32
& 0.002539
& 25.632
& 0.6788 \\
SDS (Ours)
& 64
& 32
& \textbf{0.000906}
& \textbf{25.800}
& 0.6789 \\
\bottomrule
\end{tabular*}
\end{table}

\subsection{Tile-Grouped Multi-View Accumulation for PCG}
\label{sec:tile_grouped_backward}
Tile-Grouped Multi-View Accumulation lowers the cost of assembling the sampled LM system for a multi-view batch by packing sampled rays from several views into one grouped layout, so that a single accumulation pass covers multiple views instead of one. 

Let $\mathcal{S}_t$ be the set of rays sampled at LM step $t$. 
Each ray corresponds to a sampled pixel, which lies in a tile of a training view.
Each sampled ray $i$ contributes an RGB residual block $\boldsymbol{r}_i\in\mathbb{R}^{3}$ 
and a Jacobian block $\boldsymbol{J}_i\in\mathbb{R}^{3\times P}$, 
grouping the three channel residuals $r_{i,k}$ and their derivatives for that ray.
The matrix-free product and the sampled gradient term are then given by:

\begin{equation*}
    \widehat{\boldsymbol{A}}_t\boldsymbol{u}
    =
    \sum_{i\in\mathcal{S}_t}
    \boldsymbol{J}_{i}^{\top}
    \left(
    \boldsymbol{J}_{i}\boldsymbol{u}
    \right)
    +
    \lambda\boldsymbol{u},
    \qquad
    \widehat{\boldsymbol{b}}_t
    =
    -
    \sum_{i\in\mathcal{S}_t}
    \boldsymbol{J}_{i}^{\top}
    \boldsymbol{r}_{i},
    \label{eq:tile_grouped_pcg_quantities}
\end{equation*}
for any $\boldsymbol{u}\in\mathbb{R}^P$. 
$\widehat{\boldsymbol{A}}_t$ and $\widehat{\boldsymbol{b}}_t$ are the two sides of the sampled LM system in \cref{eq:sampled_lm_system}, written as sums over the sampled rays.

Both quantities are sums over the sampled rays, so their value does not depend on how the rays are grouped. 
A naive implementation still loops over the tiles of each view, rebuilding each view's sampled rays and adding its contribution on a separate pass, so the number of passes grows with the view count. 
We instead pack sampled rays from tiles of multiple views into one grouped layout, and loop over these grouped units,
reducing the number of passes from the view count to a fraction of it while the summed result stays identical. 
The sampled rays are still defined in view-local tiles, and each view keeps its own projection, visibility order, and alpha compositing, so
grouping changes only how the rays are arranged. 
We reuse the same grouped layout for every accumulation PCG needs, namely the gradient term, the matrix-free products, and the Jacobi diagonal.

\subsection{Bucket-Prefix Jacobi Preconditioning}
\label{sec:bucket_prefix}
Bucket-Prefix Jacobi Preconditioning builds the Jacobi diagonal used for preconditioning without recomputing each pixel's compositing from the first Gaussian, by caching the transmittance and accumulated color at a few points along the front-to-back Gaussian list and resuming from them.
Besides the gradient term and the matrix-free PCG products, PCG also needs the Jacobi diagonal, which we accumulate from the same sampled rays and blocks:

\begin{equation*}
    \boldsymbol{D}_t
    =
    \operatorname{diag}
    \left(
    \sum_{i\in\mathcal{S}_t}
    \boldsymbol{J}_i^{\top}\boldsymbol{J}_i
    \right),
    \label{eq:jacobi_diagonal}
\end{equation*}

where $\boldsymbol{D}_t\in\mathbb{R}^{P\times P}$ is the Jacobi diagonal at step $t$.

Collecting this diagonal naively is expensive in 3DGS. 
For each sampled ray $i$, the required value is tied to how the pixel for that ray was rendered, and a naive kernel reruns the forward alpha-compositing pass from the first Gaussian to recover the transmittance and the color accumulated before the current Gaussian.
This recomputes the same transmittance and accumulated color over and over for the Gaussians along each ray.
Caching these two values at every Gaussian would remove the recomputation, but costs too much memory for multi-view sampled LM.

Bucket-Prefix Jacobi Preconditioning therefore caches them only at bucket boundaries.
For each sampled ray in $\mathcal{S}_t$, we split the front-to-back ordered Gaussian list of its tile into small buckets, and at each bucket boundary we store only the bucket range and two values: the transmittance and the color accumulated before that bucket.
During diagonal accumulation, all buckets start at once from their cached boundaries, each processes only the Gaussians inside it, and their contributions are summed into the diagonal. Later Gaussians thus no longer wait for the compositing of those in front of them. This avoids both the full recomputation from the first composited Gaussian along the ray and the memory cost of storing the two values at every Gaussian.

The accumulated diagonal forms a power-Jacobi preconditioner:

\begin{equation*}
    \boldsymbol{P}_t^{-1}
    =
    \left(
    \boldsymbol{D}_t
    +
    (\lambda+\epsilon)\boldsymbol{I}
    \right)^{-\beta},
    \label{eq:bucket_jacobi}
\end{equation*}

where the power is applied to each diagonal entry, $\boldsymbol{I}$ is the
identity matrix, $\lambda$ is the LM damping, $\epsilon$ prevents division by
zero, and $\beta$ sets the preconditioning strength.
This changes neither the sampled LM
system nor the matrix-free PCG operator, and only makes the diagonal cheaper to
collect.
\section{Experiments}
\label{sec:exp}

\subsection{Experimental Settings}
\label{sec:exp_settings}

% \begin{figure*}[t]
% \centering
% \includegraphics[width=\textwidth]{figures/sampling_distribution.pdf}
% \caption{
% \textbf{Effect of view selection on the rays of a batch in the Bonsai scene.}
% (a) Mismatch between the tile-center camera rays of one batch and those of all training views over a 2D projection of ray descriptors.
% Dark cells match, bright cells hold too many or too few rays, and white cells hold no training ray.
% (b) Ray Coverage Distance over 100 consecutive batches with the mean marked, where a lower value means a closer match.
% K-Means with 32 views is the setting of LM-RS, and at 64 views our batches match the rays of all training views more closely than K-Means batches (\cref{sec:exp_settings}).}
% \label{fig:sampling_distribution}
% \end{figure*}

\noindent\textbf{Datasets and Metrics.}
\label{exp_settings}
All experiments run on a single NVIDIA RTX PRO 6000 Blackwell GPU.
We evaluate on three standard 3DGS benchmarks: Mip-NeRF 360~\cite{barron2022mipnerf360}, Tanks\&Temples~\cite{knapitsch2017tanks}, and Deep Blending~\cite{hedman2018deep}, covering 13 real-world scenes with indoor and outdoor, bounded and unbounded environments.
Following the 3DGS evaluation protocol~\cite{kerbl2023gaussians}, we use the same train/test splits and report PSNR, SSIM~\cite{wang2004ssim}, and LPIPS~\cite{zhang2018lpips} averaged over all test images.
For view selection, we report Coverage Distance, which compares the camera poses of a selected batch, described by camera position and viewing direction, with those of all training views.
Lower is better.
We also report optimization time and peak GPU memory.
Additional results on Tanks\&Temples and Deep Blending, and further details of Coverage Distance, are provided in the supplementary material.

\noindent\textbf{Baselines.}
We compare against the state-of-the-art second-order methods LM-RS~\cite{lmrs} and 3DGS-LM~\cite{3dgslm}, and 3DGS as the first-order baseline.
% We follow the official implementations except for LM-RS. 
% Since the original LM-RS assumes each scene has a Lambertian surface and disables the corresponding spherical harmonics beyond degree-0, we extend LM-RS to degree-3 SH for a fair comparison.
We follow the official implementations except for LM-RS. The original LM-RS assumes each scene has a Lambertian surface and disables the corresponding spherical harmonics beyond degree-0. For a stronger baseline with better convergence, we enable degree-3 SH optimization in LM-RS and increase its original 32-view batch setting to 64 views. We refer to this strengthened setting as LM-RS 64 in our experiments. Results under the original LM-RS configuration are provided in the supplementary material.

\noindent\textbf{Optimization Protocol.}
As our goal is to improve second-order optimization, we evaluate on a fixed Gaussian set, so differences come from the optimization and not from Gaussian density control algorithms.
Therefore, all methods start from the same initial Gaussians and are compared after updating only the Gaussian parameters.

Our method follows the sampled LM update protocol of LM-RS~\cite{lmrs}: it uses residual
sampling, a matrix-free PCG solver, diagonal preconditioning, damping, and an automatic step-size rule. 
Each update optimizes the full set of standard 3DGS parameters, including degree-3 spherical harmonic coefficients (SH).
Unless otherwise stated, LM-RS with the additional SH coefficients and our method use the same PCG iteration budget, damping, view batch size, and number of updates, and differ only in how the view batch is selected and how the sampled operator and preconditioner are executed.

\begin{table*}[t]
    \centering
    \caption{
    \textbf{Optimization from SfM initialization on Mip-NeRF 360.}
    All methods start from the same SfM point cloud and optimize with the number of Gaussians fixed, Adam for 30,000 updates and the second-order methods for 300.
    Time measures optimization time.
    3DGS-LM uses a reduced $16{\times}4$ configuration to avoid running out of memory.
    Best results are in \textbf{bold} and second-best are \underline{underlined}.
    }
    \label{tab:pointcloud_init}
    \resizebox{\textwidth}{!}{
    \begin{tabular}{lcccccccc}
    \toprule
    & \multicolumn{8}{c}{Mip-NeRF 360, SfM initialization} \\
    \cmidrule(lr){2-9}
    \multirow{2}{*}{Scene type}
    & \multicolumn{2}{c}{3DGS-Adam}
    & \multicolumn{2}{c}{3DGS-LM $16{\times}4$}
    & \multicolumn{2}{c}{LM-RS 64}
    & \multicolumn{2}{c}{Ours 64} \\
    \cmidrule(lr){2-3}
    \cmidrule(lr){4-5}
    \cmidrule(lr){6-7}
    \cmidrule(lr){8-9}
    & PSNR & Time (s)
    & PSNR & Time (s)
    & PSNR & Time (s)
    & PSNR & Time (s) \\
    \midrule
    Indoor Avg.
    & 29.664 & \underline{549.5}
    & 24.219 & 1970.7
    & \underline{29.732} & 638.9
    & \textbf{29.867} & \textbf{296.1} \\
    Outdoor Avg.
    & \underline{22.525} & 383.9
    & 20.173 & 941.4
    & 22.352 & \underline{376.2}
    & \textbf{22.547} & \textbf{119.7} \\
    \midrule
    Overall Avg.
    & \underline{25.698} & \underline{457.5}
    & 21.971 & 1398.9
    & 25.632 & 492.9
    & \textbf{25.800} & \textbf{198.1} \\
    \bottomrule
    \end{tabular}
    }
\end{table*}

\begin{table*}[t]
\centering
\caption{
\textbf{Quantitative comparison of checkpoint fine-tuning on Mip-NeRF 360.}
We compare optimization methods on dense pretrained 3DGS models from
3DGS~\cite{kerbl2023gaussians},
Taming-3DGS~\cite{taming3dgs},
3DGS-MCMC~\cite{3dgsmcmc}, and
DISTWAR~\cite{durvasula2023distwar}.
Each checkpoint is the model saved at the end of densification.
Quality and time are averaged over all nine scenes.
Quality metrics report the mean $\pm$ sample standard deviation over five seeds;
Time is the mean fine-tuning time in seconds, and Mem. is the peak GPU memory
measured on Bonsai.
3DGS-LM uses a reduced $16{\times}4$ configuration that solves four 16-view subsets sequentially, so its peak memory corresponds to one 16-view solve.
Best and second-best results are in \textbf{bold} and \underline{underlined}.
}
\label{tab:checkpoint_finetuning_mip360}
\footnotesize
\setlength{\tabcolsep}{3pt}
\renewcommand{\arraystretch}{0.88}
\resizebox{\textwidth}{!}{
\begin{tabular}{lccccc}
\toprule
Fine-tuning Method (Steps)
& PSNR$\uparrow$ & SSIM$\uparrow$ & LPIPS$\downarrow$ & Time$\downarrow$ & Mem. (GB)$\downarrow$ \\
\midrule
\multicolumn{6}{l}{\textit{3DGS checkpoint}} \\
Adam (10k)
& $27.5754{\scriptstyle\pm0.0043}$
& $\mathbf{0.82571}{\scriptstyle\pm0.00020}$
& $\mathbf{0.20274}{\scriptstyle\pm0.00018}$
& $405.4$ & \textbf{9.37} \\
3DGS-LM (50)
& $27.6041{\scriptstyle\pm0.0079}$
& $\underline{0.82309}{\scriptstyle\pm0.00013}$
& $\underline{0.21301}{\scriptstyle\pm0.00006}$
& $559.8$ & 55.72 \\
LM-RS 64 (50)
& $\underline{27.8093}{\scriptstyle\pm0.0151}$
& $0.81751{\scriptstyle\pm0.00028}$
& $0.21518{\scriptstyle\pm0.00018}$
& $\underline{370.0}$ & 52.44 \\
Ours (50)
& $\mathbf{27.9097}{\scriptstyle\pm0.0009}$
& $0.81838{\scriptstyle\pm0.00004}$
& $0.21517{\scriptstyle\pm0.00004}$
& $\mathbf{177.9}$ & \underline{50.16} \\

\midrule

\multicolumn{6}{l}{\textit{Taming-3DGS checkpoint}} \\
Taming-3DGS (10k)
& $27.8931{\scriptstyle\pm0.0057}$
& $\mathbf{0.81819}{\scriptstyle\pm0.00019}$
& $\mathbf{0.22259}{\scriptstyle\pm0.00021}$
& $\underline{216.2}$ & \textbf{6.73} \\
3DGS-LM (50)
& $27.8815{\scriptstyle\pm0.0057}$
& $\underline{0.81443}{\scriptstyle\pm0.00007}$
& $\underline{0.23069}{\scriptstyle\pm0.00005}$
& $695.5$ & 63.59 \\
LM-RS 64 (50)
& $\underline{28.0256}{\scriptstyle\pm0.0122}$
& $0.80874{\scriptstyle\pm0.00022}$
& $0.23271{\scriptstyle\pm0.00015}$
& $333.7$ & 44.72 \\
Ours (50)
& $\mathbf{28.1823}{\scriptstyle\pm0.0006}$
& $0.80994{\scriptstyle\pm0.00001}$
& $0.23168{\scriptstyle\pm0.00002}$
& $\mathbf{134.9}$ & \underline{43.23} \\

\midrule

\multicolumn{6}{l}{\textit{3DGS-MCMC checkpoint}} \\
3DGS-MCMC (10k)
& $28.0141{\scriptstyle\pm0.0090}$
& $\underline{0.83885}{\scriptstyle\pm0.00014}$
& $\mathbf{0.18764}{\scriptstyle\pm0.00036}$
& $569.3$ & \textbf{8.50} \\
3DGS-LM (50)
& $28.2212{\scriptstyle\pm0.0029}$
& $\mathbf{0.84018}{\scriptstyle\pm0.00008}$
& $\underline{0.18950}{\scriptstyle\pm0.00003}$
& $686.2$ & 68.93 \\
LM-RS 64 (50)
& $\underline{28.3325}{\scriptstyle\pm0.0186}$
& $0.83508{\scriptstyle\pm0.00029}$
& $0.19247{\scriptstyle\pm0.00022}$
& $\underline{499.4}$ & 62.21 \\
Ours (50)
& $\mathbf{28.4893}{\scriptstyle\pm0.0031}$
& $0.83661{\scriptstyle\pm0.00004}$
& $0.19094{\scriptstyle\pm0.00002}$
& $\mathbf{219.9}$ & \underline{58.71} \\

\midrule

\multicolumn{6}{l}{\textit{DISTWAR checkpoint}} \\
DISTWAR (10k)
& $27.5018{\scriptstyle\pm0.0069}$
& $\mathbf{0.82561}{\scriptstyle\pm0.00018}$
& $\mathbf{0.20300}{\scriptstyle\pm0.00024}$
& $391.5$ & \textbf{8.87} \\
3DGS-LM (50)
& $27.5286{\scriptstyle\pm0.0042}$
& $\underline{0.82315}{\scriptstyle\pm0.00010}$
& $\underline{0.21261}{\scriptstyle\pm0.00002}$
& $466.8$ & \underline{48.51} \\
LM-RS 64 (50)
& $\underline{27.6774}{\scriptstyle\pm0.0132}$
& $0.81536{\scriptstyle\pm0.00030}$
& $0.21666{\scriptstyle\pm0.00020}$
& $\underline{374.2}$ & 56.84 \\
Ours (50)
& $\mathbf{27.8436}{\scriptstyle\pm0.0023}$
& $0.81709{\scriptstyle\pm0.00006}$
& $0.21591{\scriptstyle\pm0.00005}$
& $\mathbf{184.9}$ & 53.12 \\
\bottomrule
\end{tabular}
}
\end{table*}

\noindent\textbf{SfM Initialization.}
For the Mip-NeRF 360 scenes, we evaluate optimization from Gaussians initialized from the SfM point cloud following 3DGS~\cite{kerbl2023gaussians}.
For this setting, we train the first-order baseline using Adam for 30k iterations, while running second-order methods for 300 updates.
LM-RS and our method both use 64 views per batch.

\noindent\textbf{Checkpoint Fine-tuning.}
We start the optimization from dense pretrained models of 3DGS~\cite{kerbl2023gaussians}, Taming-3DGS~\cite{taming3dgs}, 3DGS-MCMC~\cite{3dgsmcmc}, and DISTWAR~\cite{durvasula2023distwar}, each saved at the end of its densification.
From each model, we continue its own optimizer for 10k iterations as the first-order reference and run each second-order method for 50 updates.

\subsection{Experimental Results}
\label{sec:exp_results}

\begin{figure*}[t]
\centering
\includegraphics[width=\textwidth]{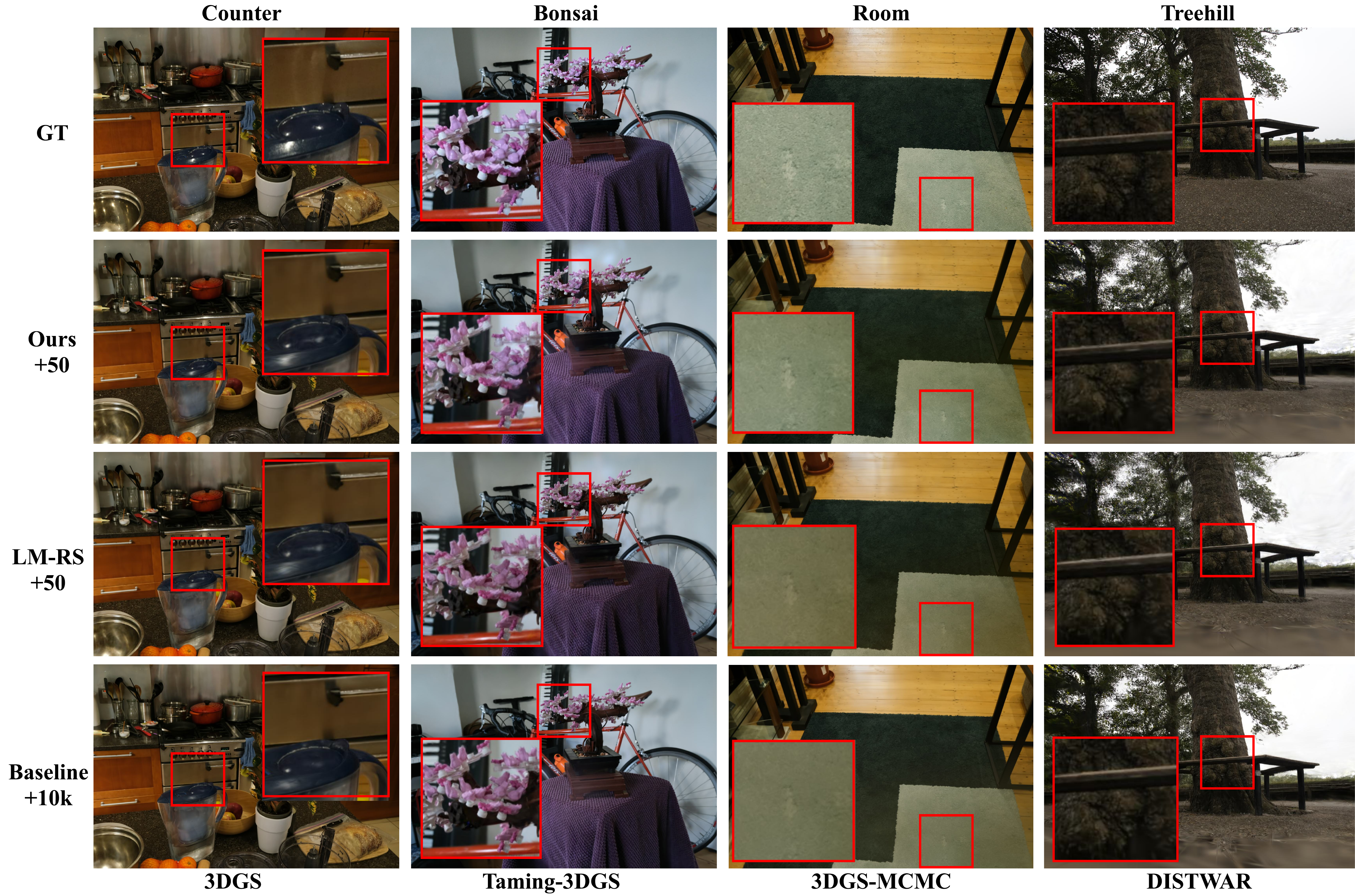}
\caption{\textbf{Qualitative comparison with baselines for fine-tuning dense pretrained 3DGS models on Mip-NeRF 360 scenes.}
Rows show the ground truth, ours and LM-RS after 50 updates, and each checkpoint's own optimizer after 10k iterations; the label under each column names the method that produced the checkpoint.
Our method better preserves view-dependent effects such as specular highlights and renders colors closer to the ground truth than LM-RS and the continued checkpoint baselines in the highlighted regions.
Large boxes enlarge the small ones.
}
\label{fig:variousbaselinequal}
\end{figure*}

\noindent\textbf{Qualitative Comparisons.}
The qualitative results in \cref{fig:variousbaselinequal} show that
our optimizer improves fine-tuning quality over the main second-order competitor
and over simply continuing each checkpoint baseline, especially in local appearance
and structural details. In the highlighted regions, our method better matches the
ground truth, including the specular reflection on the bottle in Counter, the color
and appearance of the LEGO-style bonsai in Bonsai and the carpet in Room, and thin
edge structures in Treehill.

\noindent\textbf{Effect of Scene-Distributional View Sampling.}
In \cref{tab:mip360_sampling_ablation}, we compare view-selection strategies on the nine scenes of Mip-NeRF 360 under a fixed 64-view budget.
Our sampled batches achieve lower Coverage Distance than random sampling and K-Means selection, together with the best PSNR.
This indicates that reducing redundancy in camera position and viewing direction improves the coverage of the selected batch.
A finer ray-level analysis in the supplementary material shows the same trend, although our method does not directly optimize pixel or ray-level coverage.

\noindent\textbf{Results of SfM Initialization.}
In \cref{tab:pointcloud_init}, we compare optimization from SfM initialization on the scenes of Mip-NeRF 360.
Although the 64-view LM-RS loses its speed advantage over Adam, our method reaches the highest average PSNR while running 2.5$\times$ faster than LM-RS and 2.3$\times$ faster than 3DGS-Adam, the standard 30k-iteration 3DGS training. 
This demonstrates that a well-constructed second-order update is effective not only for refinement but also for optimization from SfM initialization.

\noindent\textbf{Results of Checkpoint Fine-tuning.}
On the 3DGS checkpoint in \cref{tab:checkpoint_finetuning_mip360}, our optimizer achieves the best PSNR and the shortest fine-tuning time.
It is roughly 2$\times$ faster than LM-RS while maintaining competitive SSIM and LPIPS.
It also shows the same trend across checkpoints from Taming-3DGS, 3DGS-MCMC, and DISTWAR, where our optimizer consistently improves PSNR and reduces fine-tuning time compared with other second-order baselines.

\subsection{Ablation Study}
\label{sec:exp_ablation}

We ablate the proposed components by adding them cumulatively to the sampled LM solver, evaluated on the train and truck scenes of Tanks\&Temples in \cref{tab:ablation}.
Scene-Distributional View Sampling raises PSNR over the LM-RS 64-view baseline. 
Tile-Grouped Multi-View Accumulation then reduces the time for 300 update steps by more than 40\% while maintaining PSNR, and Bucket-Prefix Jacobi Preconditioning reduces it by a further 20\%.
Together, the three components reach higher PSNR than the sampled second-order baseline while accelerating its optimization by more than 2$\times$.

\begin{table}[t]
\centering
\caption{
\textbf{Ablation study on the train and truck scenes of Tanks\&Temples.}
Components are added cumulatively to LM-RS with the additional SH coefficients and 64 views per batch.
SDS, TG, and BP denote Scene-Distributional View Sampling,
Tile-Grouped Multi-View Accumulation, and Bucket-Prefix Jacobi Preconditioning, respectively.
PSNR and the total optimization time for the same number of updates are averaged over the two scenes, and speedup is computed relative to the baseline.
}

\label{tab:ablation}
\small
\setlength{\tabcolsep}{4pt}
\begin{tabular*}{\linewidth}{@{\extracolsep{\fill}}lccc ccc@{}}
\toprule
Method & SDS & TG & BP & PSNR$\uparrow$ & Time (s)$\downarrow$ & Speedup$\uparrow$ \\
\midrule
LM-RS 64 Views per Batch
& \xmark & \xmark & \xmark & 23.481 & 255.2 & 1.00$\times$ \\
+ SDS
& \cmark & \xmark & \xmark & 23.739 & 245.5 & 1.04$\times$ \\
+ SDS + TG
& \cmark & \cmark & \xmark & 23.742 & 144.6 & 1.76$\times$ \\
+ SDS + TG + BP (Ours)
& \cmark & \cmark & \cmark & 23.742 & \textbf{113.9} & \textbf{2.24}$\times$ \\
\bottomrule
\end{tabular*}
\end{table}

\section{Conclusion}
\label{sec:con}

Our results show that second-order optimization becomes effective for 3DGS when the multi-view batch is both scene-representative and efficient to execute.
Scene-Distributional View Sampling selects multi-view batches with less redundancy, so that each sampled second-order update draws on views whose camera poses are distributed more like those of all training views.
Tile-Grouped Multi-View Accumulation and Bucket-Prefix Jacobi Preconditioning make these sampled second-order updates practical by shortening the accumulation of the sampled LM system and the computation of its preconditioner.
Together, these components let our method reach higher reconstruction quality than the sampled second-order baseline, and higher PSNR than Adam in fewer updates and less time, both from sparse SfM point clouds and from dense pretrained 3DGS models.

\noindent\textbf{Limitations and Future Work.}
While our grouping and caching substantially reduce the cost of each multi-view second-order update, they cannot fully eliminate the memory overhead inherent to multi-view batch training. Because each Gaussian is ordered differently from each camera, every view keeps its own visibility order, transmittance, and accumulated colors, and this per-view state remains the dominant memory cost, limiting the batch sizes practical on commodity hardware. A shared, view-independent representation of visibility and compositing could keep memory from scaling with the number of views.

\subsubsection*{Acknowledgments.}
% This work was supported by the National Research Foundation of Korea (RS-2026-25497603: Fundamentals of Online 4D Reconstruction with Semantic Viewpoint Control and Generative Priors for Proactive Content Consumption) and by the IITP grant [RS-2020-II201361, Artificial Intelligence Graduate School Program (Yonsei University)] funded by the Korean government (MSIT).
% This work was supported by the National Research Foundation of Korea (RS-2026-25497603) and the IITP grant (RS-2020-II201361), funded by the Korean government (MSIT).
This work was supported by National Research Foundation of Korea (RS-2026-25497603), IITP (RS-2020-II201361), and the “Advanced GPU Utilization Support Program”, funded by the Korean government (MSIT).

% \nocite{supple에만 있는 논문 키}
\bibliographystyle{splncs04}
\bibliography{main}

% \clearpage
% \input{supple_sec/supplementary}

\end{document}